\documentclass{llncs}

\usepackage{pdflscape}
\usepackage{url}
\usepackage{verbatim}
\usepackage{eurosym}
\usepackage{amssymb}
\usepackage{graphicx}
\usepackage{multirow}
\usepackage{pifont}

\usepackage{listings}
\usepackage{framed}
\usepackage[listings,skins]{tcolorbox}
\usepackage[skipbelow=\topskip,skipabove=\topskip]{mdframed}
\mdfsetup{roundcorner=0}
\usepackage{etoolbox}

\usepackage{verbatim}

\usepackage{algorithmicx}
\usepackage{algorithm}
\usepackage{algpseudocode}
\algnewcommand\algorithmicinput{\textbf{Input:}}
\algnewcommand\INPUT{\item[\algorithmicinput]}

\algnewcommand\algorithmicoutput{\textbf{Output:}}
\algnewcommand\OUTPUT{\item[\algorithmicoutput]}

\usepackage{url}
\begin{document}

\title{AVATAR - Machine Learning Pipeline Evaluation Using Surrogate Model}

\author{Tien-Dung Nguyen$^{1}$ \and
Tomasz Maszczyk$^{1}$ \and 
Katarzyna Musial$^{1}$ \and \\
Marc-Andr\'e Z\"oller$^{2}$ \and 
Bogdan Gabrys$^{1}$}

\institute{$^{1}$University of Technology Sydney, Sydney, Australia 
\email{\{TienDung.Nguyen-2@student.,Tomasz.Maszczyk@,\\Katarzyna.Musial-Gabrys@,Bogdan.Gabrys@\}uts.edu.au} \\
$^{2}$USU Software AG, Karlsruhe, Germany  \\
\email{m.zoeller@usu.de}}
\maketitle
\begin{abstract}

The evaluation of machine learning (ML) pipelines is essential during automatic ML pipeline composition and optimisation. The previous methods such as Bayesian-based and genetic-based optimisation, which are implemented in Auto-Weka, Auto-sklearn and TPOT, evaluate pipelines by executing them. Therefore, the pipeline composition and optimisation of these methods requires a tremendous amount of time that prevents them from exploring complex pipelines to find better predictive models. 
To further explore this research challenge, we have conducted experiments showing that many of the generated pipelines are invalid, and it is unnecessary to execute them to find out whether they are good pipelines.    
To address this issue, we propose a novel method to evaluate the validity of ML pipelines using a surrogate model (AVATAR). The AVATAR enables to accelerate automatic ML pipeline composition and optimisation by quickly ignoring invalid pipelines. Our experiments show that the AVATAR is more efficient in evaluating complex pipelines in comparison with the traditional evaluation approaches requiring their execution.
\end{abstract}

\section{Introduction}
\label{sec:intro}

Automatic machine learning (AutoML) has been studied to automate the process of data analytics to collect and integrate data, compose and optimise ML pipelines, and deploy and maintain predictive models \cite{kaga09,sabu18,zohu19}. Although many existing studies proposed methods to tackle the problem of pipeline composition and optimisation \cite{sabu18,olmo16,fekl15,mowe18,giya18,depi17,tsga12}, these methods have two main drawbacks.
Firstly, the pipelines' structures, which define the executed order of the pipeline components, use fixed templates \cite{sabu18,fekl15}. Although using fixed structures can reduce the number of invalid pipelines during the composition and optimisation, these approaches limit the exploration of promising pipelines which may have a variety of structures.
Secondly, while evolutionary algorithms based methods \cite{olmo16} enable the randomness of the pipelines' structure using the concept of evolution, this randomness tends to construct more invalid pipelines than valid ones. Besides, the search spaces of the pipelines' structures and hyperparameters of the pipelines' components expand significantly. Therefore, the existing approaches tend to be inefficient as they often attempt to evaluate invalid pipelines. 
There are several attempts to reduce the randomness of pipeline construction by using context-free grammars \cite{depi17,tsga12} or AI planning to guide the construction of pipelines \cite{mowe18,giya18}. Nevertheless, all of these methods evaluate the validity of a pipeline by executing them (T-method). After executing a pipeline, if the result is a predictive model, the T-method evaluates the pipeline to be valid; otherwise it is invalid. If a pipeline is complex, the complexity of preprocessing/predictor components within the pipeline is high, or the size of the dataset is large, the evaluation of the pipeline is expensive. Consequently, the optimisation will require a significant time budget to find well-performing pipelines.

To address this issue, we propose the AVATAR to evaluate ML pipelines using their surrogate models. The AVATAR transforms a pipeline to its surrogate model and evaluates it instead of executing the original pipeline.
We use the business process model and notation (BPMN)\cite{chmi09} to represent ML pipelines.
BPMN was invented for the purposes of a graphical representation of business processes, as well as a description of resources for process execution. In addition, BPMN  simplifies the understanding of business activities and interpretation of behaviours of ML pipelines. The ML pipelines' components use the Weka libraries\footnote{\url{https://www.cs.waikato.ac.nz/ml/weka/}} for ML algorithms.
The evaluation of the surrogate models requires a knowledge base which is generated from many synthetic datasets.
To this end, this paper has two main contributions:

\begin{itemize}
    \item We conduct experiments on current state-of-the-art AutoML tools to show that the construction of invalid pipelines during the pipeline composition and optimisation may lead to bad performance.

    \item We propose the AVATAR to accelerate the automatic pipeline composition and optimisation by evaluating pipelines using a surrogate model. 
\end{itemize}

This paper is divided into five sections. After the Introduction, Section \ref{sec:related_work} reviews previous approaches to representing and evaluating ML pipelines in the context of AutoML.
Section \ref{sec:avatar_method} presents the AVATAR to evaluate ML pipelines.
Section \ref{sec:experiment} presents experiments to motivate our research and prove the efficiency of the proposed method. Finally, Section \ref{sec:conclusion} concludes this study.

\section{Related Work}
\label{sec:related_work}

%This section aims to review formal pipeline representations and pipeline evaluation methods of the current state-of-the-art pipeline composition and optimisation methods which are implemented as the AutoML tools such as AutoWeka for multi-component predictive systems (AutoWeka4MCPS) \cite{sabu18}, ML-Plan \cite{mowe18}, P4ML \cite{giya18}, TPOT \cite{olmo16} and Auto-sklearn \cite{fekl15}.

Salvador et al. \cite{sabu18} proposed an automatic pipeline composition and optimisation method of multicomponent predictive systems (MCPS) to deal with the problem of combined algorithm selection and hyperparameter optimisation (CASH). This proposed method is implemented in the tool AutoWeka4MCPS \cite{sabu18} developed on top of Auto-Weka 0.5 \cite{thhu13}. The pipelines, which are generated by AutoWeka4MCPS, are represented using Petri nets \cite{sabu17b}.
A Petri net is a mathematical modelling language used to represent pipelines \cite{sabu18} as well as data service compositions \cite{tafa10}.
The main idea of Petri nets is to represent transitions of states of a system. 
Although it is not clearly mentioned in these previous works \cite{olmo16,fekl15,mowe18,giya18}, directed acyclic graph (DAG) is often used to model sequential pipelines in the methods/tools such as AutoWeka4MCPS \cite{sabu16b}, ML-Plan \cite{mowe18}, P4ML \cite{giya18}, TPOT \cite{olmo16} and Auto-sklearn \cite{fekl15}.
DAG is a type of graph that has connected vertexes, and the connections of vertexes have only one direction \cite{bava07}. In addition, a DAG does not allow any directed loop. It means that it is a topological ordering. 
ML-Plan generates sequential workflows consisting of ML components. Thus, the workflows are a type of DAG. The final output of P4ML is a pipeline which is constructed by making an ensemble of other pipelines. Auto-sklearn generates fixed-length sequential pipelines consisting of scikit-learn components. TPOT construct pipelines consisting of multiple preprocessing sub-pipelines. The authors claim that the representation of the pipelines is a tree-based structure. However, a tree-based structure always starts with a root node and ends with many leaf nodes, but the output of a TPOT's pipeline is a single predictive model. Therefore, the representation of TPOT pipeline is more like a DAG. P4ML uses a tree-based structure to make a multi-layer ensemble. This tree-based structure can be specialised into a DAG. The reason is that the execution of these pipelines will start from leaf nodes and end at root nodes where the construction of the ensembles are completed. It means that the control flows of these pipelines have one direction, or they are topologically ordered. 
Using a DAG to model an ML pipeline makes it easy to understand by humans as DAGs facilitate visualisation and interpretation of the control flow. However, DAGs do not model inputs/outputs (i.e. possibly datasets, output predictive models, parameters and hyperparameters of components) between vertexes. Therefore, the existing studies use ad-hoc approaches and make assumptions about data inputs/outputs of the pipelines' components. 
%BPMN is the most promising method to represent an ML pipeline. The reasons are that a BPMN-based ML pipeline can be executable, has a better interpretation of the pipeline in terms of control, data flows and resources for execution, as well as integrates into existing business processes as a subprocess. 
%Moreover, we claim that a Petri Net is the most promising method to represent a surrogate pipeline. The reason is that it is fast to verify the validity of a Petri-net based simplified ML pipeline.

Although AutoWeka4MCPS, ML-Plan, P4ML, TPOT and Auto-sklearn evaluate pipelines by executing them, these methods have strategies to limit the generation of invalid pipelines. Auto-sklearn uses a fixed pipeline template including preprocessing, predictor and ensemble components.
%TPOT evaluates a pipeline on a sampling dataset which has the first five instances of its original dataset.
AutoWeka4MCPS also uses a fixed pipeline template consisting of six components. TPOT, ML-Plan and P4ML use grammars/ primitive catalogues, which are designed manually, to guide the construction of pipelines. Although these approaches can reduce the number of invalid pipelines, our experiments showed that the wasted time used to evaluate the invalid pipelines is significant. Moreover, using fixed templates, grammars and primitive catalogues reduce search spaces of potential pipelines, which is a drawback during pipeline composition and optimisation. 
\section{Evaluation of ML Pipelines Using Surrogate Models}
\label{sec:avatar_method}

%\subsection{knowledge base for Constructing Machine Learning Pipelines}
%\label{sec:meta_knowledge}

Because the evaluation of ML pipelines is expensive in certain cases (i.e., complex pipelines, high complexity pipeline's components and large datasets) in the context of AutoML, we propose the AVATAR\footnote{\url{https://github.com/UTS-AAi/AVATAR}} to speed up the process by evaluating their surrogate pipelines.
The main idea of the AVATAR is to expand the purpose and representation of MCPS introduced in \cite{sabu17b}. The AVATAR uses a surrogate model in the form of a Petri net. This surrogate pipeline keeps the structure of the original pipeline, replaces the datasets in the form of data matrices (i.e., components' input/output simplified mappings) by the matrices of transformed-features, and the ML algorithms by transition functions to calculate the output from the input tokens (i.e., the matrices of transformed-features). Because of the simplicity of the surrogate pipelines in terms of the size of the tokens and the simplicity of the transition functions, the evaluation of these pipelines is substantially less expensive than the original ones.
%The implementation of AVATAR can be found here\footnote{\url{https://github.com/UTS-AAi/AVATAR}}.

%\subsection{Transformed-features}
%\label{subsec:transformed_features}

%We classify these transformed-features into three types as follows.

% \begin{itemize}
%     \item Quality: The quality type transformed-features (e.g., \textit{MISSING\_VALUES}) describe quality-related characteristics of a dataset. Basically, filters can be used to improve these quality transformed-features.   
%     \item Attribute Transformation: The attribute transformation type transformed-features (e.g., \textit{NUMERIC\_ATTRIBUTES}) describe a dataset's characteristics which can be transformed into another attribute transformation type using a filter.
%     \item Class Transformation: This type of transformed-feature is similar to attribute transformation. The only difference is that it is applicable for classes.
% \end{itemize}

\subsection{The AVATAR knowledge base}
\label{subsec:metaknowledge}

\begin{table}[htb!]
\centering

\vspace{-0.35cm}
\caption {Descriptions of the transformed-features of a dataset.}
\label{tab:transformed_features}

\centering

\scriptsize
\begin{tabular}{|l|l|}
\hline
\textbf{Transformed-feature}                                          & \textbf{Description}                        \\ \hline
BINARY\_CLASS                                                         & a dataset has binary classes                \\ \hline
NUMERIC\_CLASS                                                         & a dataset has numeric classes                \\ \hline
DATE\_CLASS                                                           & a dataset has date classes                  \\ \hline
MISSING\_CLASS\_VALUES                                                & a dataset has missing values in classes     \\ \hline
NOMINAL\_CLASS                                                        & a dataset has nominal classes               \\ \hline
SYMBOLIC\_CLASS                                                       & a dataset has symbolic data in classes      \\ \hline
STRING\_CLASS                                                         & a dataset has string classes                \\ \hline
UNARY\_CLASS                                                          & a dataset has unary classes                 \\ \hline
BINARY\_ATTRIBUTES                                                    & a dataset has binary attributes             \\ \hline
DATE\_ATTRIBUTES                                                      & a dataset has date attributes               \\ \hline
\begin{tabular}[c]{@{}l@{}}EMPTY\_NOMINAL\_\\ ATTRIBUTES\end{tabular} & a dataset has an empty column               \\ \hline
MISSING\_VALUES                                                       & a dataset has missing values in attributes  \\ \hline
NOMINAL\_ATTRIBUTES                                                   & a dataset has nominal attributes            \\ \hline
NUMERIC\_ATTRIBUTES                                                   & a dataset has numeric attributes            \\ \hline
UNARY\_ATTRIBUTES                                                     & a dataset has unary attributes              \\ \hline
PREDICTIVE\_MODEL                                                     & a predictive model generated by a predictor \\ \hline
\end{tabular}
\vspace{-0.35cm}

\end{table}

% \begin{figure}[htb]
% \centering
% \includegraphics[width=1\linewidth]{images/metaknowledge_explain.pdf}
% \vspace{-0.8cm}
% \caption{An illustration of the capabilities and effects of a machine learning algorithm}
% \label{fig:meta_knowledge}

% \end{figure}

%, as illustrated in Figure \ref{fig:meta_knowledge}.
%Therefore, this knowledge base of machine learning algorithms needs to describe both these algorithms' capabilities and effects. The elements of the capabilities and effects are transformed-features.
We define transformed-features as the features, which represent dataset's characteristics. These characteristics can be changed because of the transformations of this dataset by ML algorithms.
Table \ref{tab:transformed_features} describes the transformed-features used for the knowledge base. We select these transformed-features because the capabilities of a ML algorithm to work with a dataset depend on these transformed-features. These transformed-features are extended from the capabilities of Weka algorithms\footnote{\url{http://weka.sourceforge.net/doc.dev/weka/core/Capabilities.html}}. 

The purpose of the AVATAR knowledge base is for describing the logic of transition functions of the surrogate pipelines. The logic includes the capabilities and effects of ML algorithms (i.e., pipeline components).

The capabilities are used to verify whether an algorithm is compatible to work with a dataset or not. For example, whether the linear regression algorithm can work with missing value and numeric attributes or not? The capabilities have a list of transformed-features. The value of each capability-related transformed-feature is either 0 (i.e., the algorithm can not work with the dataset which has this transformed-feature) or 1 (i.e., the algorithm can work with the dataset which has this transformed-feature).
Based on the capabilities, we can determine which components of a pipeline (i.e., ML algorithms) are not able to process specific transformed-features of a dataset. 

The effects describe data transformations. Similar to the capabilities, the effects have a list of transformed-features. Each effect-related transformed-feature can have three values, 0 (i.e., do not transform this transformed-feature), 1 (i.e., transform one or more attributes/classes to this transformed-feature), or -1 (i.e., disable the effect of this transformed-feature on one or more attributes/classes).

%For example, \textit{EMImputation} algorithm fills in all missing values of attributes; therefore we set \textit{MISSING\_VALUES = 1} in the effects.
%For another example, \textit{Component Analysis} algorithm transforms nominal attributes to numeric attributes; therefore we set \textit{NUMERIC}\textit{\_ATTRIBUTES} \textit{= 1} and \textit{NOMINAL} \textit{\_ATTRIBUTES} \textit{= -1} in the effects.

%\subsection{Synthetic Datasets for Generating the knowledge base}
%\label{sec:synthetic_dataset}

%Table \ref{tab:dataset} summarises characteristics of synthetic datasets which we use to generate the knowledge base.
%Each dataset has at least one transformed-feature described in Table \ref{tab:meta_features}. 

%\input{tables/table_dataset.tex}

%\subsection{Algorithm for generating the AVATAR knowledge base}
\label{sec:algorithm_generating_metaknowledge}

\begin{figure}[htb!]
\centering
\includegraphics[width=1.0\linewidth]{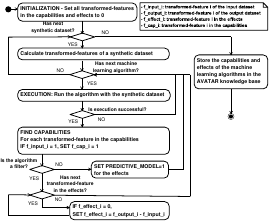}
\vspace{-0.3cm}
\caption{Algorithm to generate the knowledge base for evaluating surrogate pipelines.}
\label{fig:generate_meta_knowledge}
\end{figure}

To generate the AVATAR knowledge base\footnote{\url{https://github.com/UTS-AAi/AVATAR/blob/master/avatar-knowledge-base/avatar_knowledge_base.json}}, we have to use synthetic datasets\footnote{\url{https://github.com/UTS-AAi/AVATAR/tree/master/synthetic-datasets}} to minimise the number of active transformed-features in each dataset to evaluate which and how transformed-features impact on the capabilities and effects of ML algorithms\footnote{\url{https://github.com/UTS-AAi/AVATAR/blob/master/supplementary-documents/avatar_algorithms.txt}}. Real-world datasets usually have many active transformed-features that make them not suitable for our purpose. We minimise the number of available transformed-features in each synthetic dataset so that the knowledge base can be applicable in a variety of pipelines and datasets. Figure \ref{fig:generate_meta_knowledge} presents the algorithm to generate the AVATAR knowledge base. This algorithm has four main stages:

\begin{enumerate}

    \item Initialisation: The first stage initialises all transformed-features in the capabilities and effects to 0.
    \item Execution: Run ML algorithms with every synthetic dataset and get outputs (i.e., output datasets or predictive models).
    \item Find capabilities: If the execution is successful, we set the active transformed-features of the input dataset for the ones in the capabilities.
    \item Find effects: If an algorithm is a predictor/transformed-predictor, we set \textit{PREDICTIVE\_MODEL} for its effects. If the algorithm is a filter and its current value is a default value, we set this effect-related transformed-feature equal the difference of the values of this transformed-feature of the output and input dataset. 
\end{enumerate}

\subsection{Evaluation of ML pipelines}

The AVATAR evaluates a ML pipeline by mapping it to its surrogate pipeline and evaluating this surrogate pipeline. BPMN is the most promising method to represent an ML pipeline. The reasons are that a BPMN-based ML pipeline can be executable, has a better interpretation of the pipeline in terms of control, data flows and resources for execution, as well as integrates into existing business processes as a subprocess. 
Moreover, we claim that a Petri net is the most promising method to represent a surrogate pipeline. The reason is that it is fast to verify the validity of a Petri net based simplified ML pipeline.

\begin{figure}[htb!]
\centering
\includegraphics[width=0.95\linewidth]{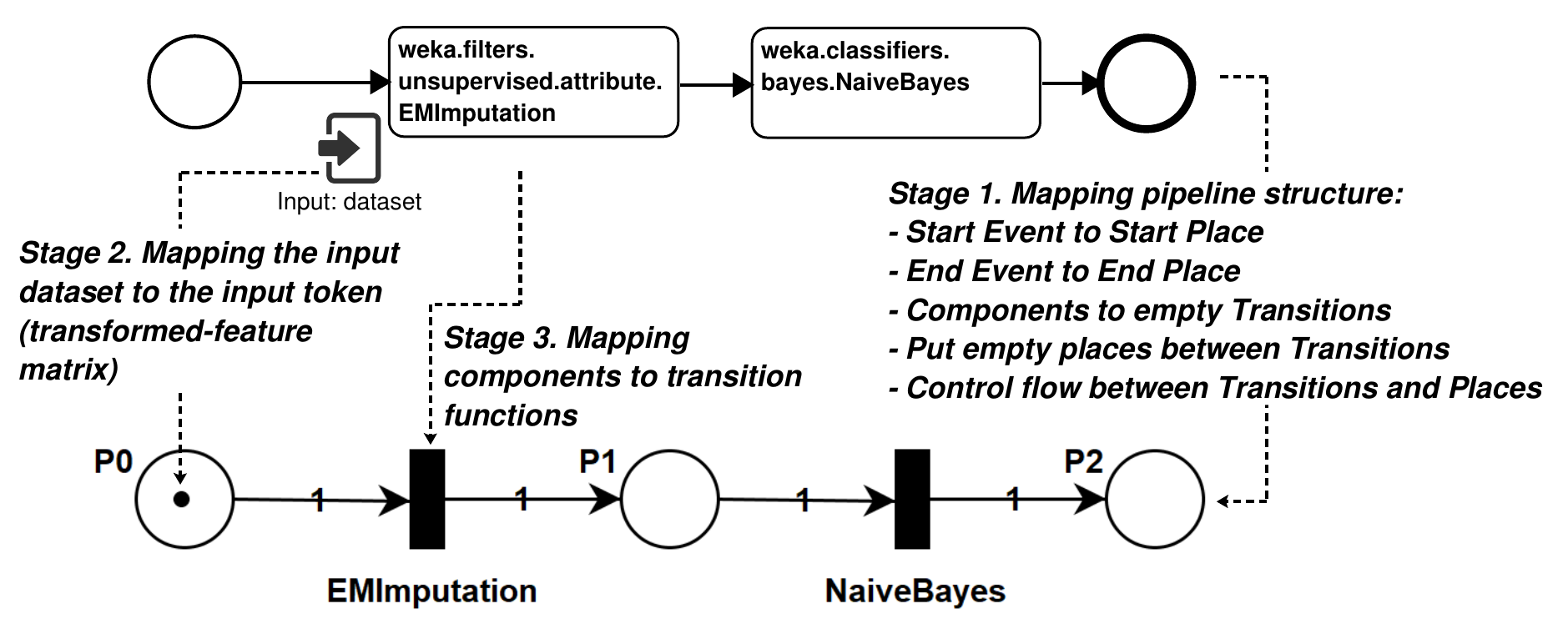}
\vspace{-0.3cm}
\caption{Mapping a ML pipeline to its surrogate model.}
\label{fig:mapping_surrogate}
\end{figure}

\subsubsection{Mapping a ML pipeline to its surrogate model}

The AVATAR maps a BPMN pipeline to a Petri net pipeline via three stages (Figure \ref{fig:mapping_surrogate}).

\begin{enumerate}

\item The structure of the BPMN-based ML pipeline is mapped to the respective structure of the Petri net surrogate pipeline. The start and end events are mapped to the start and end places respectively. The components are mapped to empty transitions. Empty places are put between all transitions. Finally, all flows are mapped to arcs.

\item The values of transformed-features are calculated from the input dataset to form a transformed-feature matrix which is the input token in the start place of the surrogate pipeline.

\item The transition functions are mapped from the components. In this stage, only the corresponding algorithm information is mapped to the transition function.    
\end{enumerate}

\subsubsection{Evaluating a surrogate model}

\begin{figure}[htb]
\centering
\includegraphics[width=1.0\linewidth]{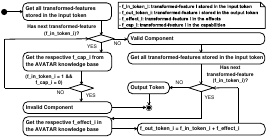}
\vspace{-0.3cm}
\caption{Algorithm for firing a transition of the surrogate model.}
\label{fig:alg_transition}
\end{figure}
The evaluation of a surrogate model will execute a Petri net pipeline. This execution starts by firing each transition of the Petri net pipeline and transforming the input token.
As shown in Figure \ref{fig:alg_transition}, firing a transition consists of two tasks: (i) the evaluation of the capabilities of each component; and (ii) the calculation of the output token.
The first task verifies the validity of the component using the following rules. If the value of a transformed-feature stored in the input token ($f\_in\_token\_i$) is 1 and the corresponding transformed-feature in the component's capabilities ($f\_cap\_i$) is 0, this component is invalid. Otherwise, this component is always valid. If a component is invalid, the surrogate pipeline is evaluated as invalid.
%For example, if the input dataset has missing values in class (i.e., MISSING\_CLASS\_VALUES=1) and the component can handle missing values in class (i.e., MISSING\_CLASS\_VALUES=1), the component is valid. 
%If the input dataset has missing values in class (i.e., MISSING\_CLASS\_VALUES=1) and the algorithm of the component cannot handle missing values in class (i.e., MISSING\_CLASS\_VALUES=0), the component is invalid. 
%If the input dataset does not have missing values in class (i.e., MISSING\_CLASS\_VALUES=0), the component is always valid.
The second task calculates each transformed-feature stored in the output token (\textit{f\_out\_token\_i}) in the next place from the input token by adding the value of a transformed-feature stored in the input token (\textit{f\_in\_token\_i}) and the respective transformed-feature in the component's effects (\textit{f\_effect\_i}).

%It is assumed that an effect-related transformed-feature $X$ is MISSING\_CLASS\_VALUES. 
%For example, if the input dataset has missing values in class (i.e., $MISSING\_CLASS\_VALUES = 1$) and the component can remove/replace missing values in class (i.e., $MISSING\_CLASS\_VALUES = -1$), so that the output dataset will not have missing values in class.
%(i.e., the transformed-feature of the output token is 1-1=0).

%If the input dataset has missing values in class (i.e., the transformed-feature X of the input token is 1) and the algorithm of the component cannot remove/replace missing values in class (i.e., $MISSING\_CLASS\_VALUES = 0$), the output dataset still has missing values in class (i.e., the transformed-feature X of the output token is 1+0=0). If the input dataset has no missing values in class (i.e., the transformed-feature X of the input token is 0), the output dataset also has no missing value in class (i.e., 0-1=-1). In this case, because the values of transformed-features of a dataset are either 0 or 1, the minimum value 0 is set to $MISSING\_CLASS\_VALUES$.

\section{Experiments}
\label{sec:experiment}

To investigate the impact of invalid pipelines on ML pipeline composition and optimisation, we have first conducted a series of experiments with current state-of-the-art AutoML tools. After that, we have conducted the experiments to compare the performance of the AVATAR and the existing methods.

\subsection{Experimental settings}

\begin{table}[htb!]

\vspace{-0.35cm}
\caption {Summary of datasets' characteristics: the number of numeric attributes, nominal attributes, distinct classes, instances in training and testing sets.}
\label{tab:exp_datasets}

\centering
\scriptsize
\begin{tabular}{|l|l|l|l|l|l|}
\hline
\textbf{Dataset} & \textbf{Numeric} & \textbf{Nominal} & \textbf{No of distinct classes} & \textbf{Training} & \textbf{Testing} \\ \hline
abalone          & 7                & 1                & 26                              & 2,924          & 1,253         \\ \hline
car              & 0                & 6                & 4                               & 1,210          & 518           \\ \hline
convex           & 784              & 0                & 2                               & 8,000          & 50,000        \\ \hline
gcredit          & 7                & 13               & 2                               & 700            & 300           \\ \hline
wineqw           & 11               & 0                & 7                               & 3,429          & 1,469         \\ \hline
\end{tabular}

\vspace{-0.35cm}

\end{table}

Table \ref{tab:exp_datasets} summarises characteristics of datasets\footnote{\url{https://archive.ics.uci.edu}} used for experiments. We use these datasets because they were used in previous studies \cite{sabu18,olmo16,fekl15}. The AutoML tools used for the experiments are AutoWeka4MCPS \cite{sabu18} and Auto-sklearn \cite{fekl15}. These tools are selected because their abilities to construct and optimise hyperparameters of complex ML pipelines have been empirically proven to be effective in a number of previous studies \cite{sabu18,fekl15,baal18}. However, these previous experiments had not investigated the negative impact of the evaluation of invalid pipelines on the quality of the pipeline composition and optimisation yet. This is the goal of our first set of experiments. 
In the second set of experiments, we show that the AVATAR can significantly reduce the evaluation time of ML pipelines.
%As a result, the number of pipelines generated during the pipeline composition and optimisation are higher than the existing methods.

\subsection{Experiments to investigate the impact of invalid pipelines}

\begin{table}[htb!]

\centering

\vspace{-0.35cm}
\caption {Negative impacts of invalid pipelines in pipeline composition and optimisation of AutoWeka4MCPS. (1): the number of invalid/ valid pipelines, (2): the total evaluation time of invalid/ valid pipelines (s), (3): the wasted evaluation time (\%).}
\label{tab:neg_impact_invalid_pipeline_aw}

\scriptsize
\begin{tabular}{|l|l|l|l|l|l|l|}
\hline
\textbf{Dataset}                  & \textbf{Criteria} & \textbf{Iter 1} & \textbf{Iter 2} & \textbf{Iter 3}          & \textbf{Iter 4}          & \textbf{Iter 5}          \\ \hline
\multirow{3}{*}{\textbf{abalone}} & \textbf{(1)}      & 16/26           & 90/79           & 69/88                    & 34/29                    & 53/80                    \\ \cline{2-7} 
                                  & \textbf{(2)}      & 3607.7/1322.5   & 2007.1/1236.4   & 4512.9/2172.3            & 3615.4/277.6             & 23.2/3509.0              \\ \cline{2-7} 
                                  & \textbf{(3)}      & 73.18           & 61.88           & 67.51                    & 92.87                    & 0.66                     \\ \hline
\multirow{3}{*}{\textbf{car}}     & \textbf{(1)}      & 205/152         & 108/70          & 197/313                  & 139/156                  & 85/64                    \\ \cline{2-7} 
                                  & \textbf{(2)}      & 3818.1/291.8    & 3498.5/113.0    & 4523.6/532.6             & 5232.2/251.3             & 4365.1/90.1              \\ \cline{2-7} 
                                  & \textbf{(3)}      & 92.90           & 96.87           & 89.47                    & 95.42                    & 97.98                    \\ \hline
\multirow{3}{*}{\textbf{convex}}  & \textbf{(1)}      & 18/20           & 2/0             & 17/11                    & \multirow{3}{*}{crashed} & \multirow{3}{*}{crashed} \\ \cline{2-5}
                                  & \textbf{(2)}      & 76.3/3588.1     & 3475.2/0.0      & 1324.7/2331.8            &                          &                          \\ \cline{2-5}
                                  & \textbf{(3)}      & 2.08            & 100.00          & 36.23                    &                          &                          \\ \hline
\multirow{3}{*}{\textbf{gcredit}} & \textbf{(1)}      & 112/195         & 229/364         & 208/166                  & 12/54                    & 30/54                    \\ \cline{2-7} 
                                  & \textbf{(2)}      & 2821.0/2260.1   & 3829.8/285.6    & 3933.8/184.0             & 3667.6/34.1              & 3634.8/64.7              \\ \cline{2-7} 
                                  & \textbf{(3)}      & 55.52           & 93.06           & 95.53                    & 99.08                    & 98.25                    \\ \hline
\multirow{3}{*}{\textbf{wineqw}}  & \textbf{(1)}      & 203/213         & 121/139         & \multirow{3}{*}{crashed} & 201/302                  & 36/54                    \\ \cline{2-4} \cline{6-7} 
                                  & \textbf{(2)}      & 4880.6/1052.9   & 4183.4/1078.6   &                          & 2418.5/1132.2            & 1639.2/862.2             \\ \cline{2-4} \cline{6-7} 
                                  & \textbf{(3)}      & 82.26           & 79.50           &                          & 68.11                    & 65.53                    \\ \hline
\end{tabular}
%\vspace{-0.35cm}
\end{table}

\begin{table}[htb!]

\centering

%\vspace{-0.35cm}

\caption {Negative impacts of invalid pipelines in pipeline composition and optimisation of Auto-sklearn. (1): the number of invalid/ valid pipelines, (2): the total evaluation time of invalid/ valid pipelines (s), (3): the wasted evaluation time (\%).}
\label{tab:neg_impact_invalid_pipeline_autosklearn}
\scriptsize
\begin{tabular}{|l|l|l|l|l|l|l|}
\hline
\textbf{Dataset}                 & \textbf{Criteria} & \textbf{Iter 1} & \textbf{Iter 2} & \textbf{Iter 3} & \textbf{Iter 4} & \textbf{Iter 5} \\ \hline
\textbf{abalone}                 & \textbf{}         & crashed         & crashed         & crashed         & crashed         & crashed         \\ \hline
\textbf{car}                     & \textbf{}         & crashed         & crashed         & crashed         & crashed         & crashed         \\ \hline
\multirow{3}{*}{\textbf{convex}} & \textbf{(1)}      & 2/13            & 2/6             & 2/8             & 2/6             & 2/8             \\ \cline{2-7} 
                                 & \textbf{(2)}      & 560.8/2981.8    & 537.7/629.2     & 584.1/1537.5    & 558.1/977.1     & 560.0/1655.9    \\ \cline{2-7} 
                                 & \textbf{(3)}      & 15.76           & 15.07           & 16.39           & 15.66           & 15.72           \\ \hline
\textbf{gcredit}                 & \textbf{}         & crashed         & crashed         & crashed         & crashed         & crashed         \\ \hline
\multirow{3}{*}{\textbf{wineqw}} & \textbf{(1)}      & 0/42            & 0/22            & 0/42            & 0/32            & 0/32            \\ \cline{2-7} 
                                 & \textbf{(2)}      & 0.0/3523.4      & 0.0/909.7       & 0.0/3197.4      & 0.0/3054.0      & 0.0/3163.5      \\ \cline{2-7} 
                                 & \textbf{(3)}      & 0.00            & 0.00            & 0.00            & 0.00            & 0.00            \\ \hline
\end{tabular}
\vspace{-0.35cm}

\end{table}

To investigate the impact of invalid pipelines, we use five iterations (Iter) for the first set of experiments. We run these experiments on AWS EC2 $t3a.small$ virtual machines which have 2 vCPU and 2GB memory. Each iteration uses a different seed number.
We set the time budget to 1 hour and the memory to 1GB.
We evaluate the pipelines produced by the AutoML tools using three criteria: (1) the number of invalid/valid pipelines, (2) the total evaluation time of invalid/ valid pipelines (seconds), and (3) the wasted evaluation time (\%). The wasted evaluation time is calculated by the percentage of the total evaluation time of invalid pipelines over the total runtime of the pipeline composition and optimisation. The wasted evaluation time represents the degree of negative impacts of invalid pipelines.
%on the pipeline composition and optimisation.
%are different among these AutoML tools. 

Table \ref{tab:neg_impact_invalid_pipeline_aw} and  \ref{tab:neg_impact_invalid_pipeline_autosklearn} present negative impacts of invalid pipelines in ML pipeline composition and optimisation of AutoWeka4MCPS and Auto-sklearn using the above criteria.
%Please refer to the supplementary document for more details on the experimental results\footnote{\url{https://github.com/UTS-AAi/AVATAR/blob/master/supplementary-documents/avatar-supplementary-doc.pdf}}. 
These tables show that not all of constructed pipelines are valid.
%Moreover, initialisation of invalid pipelines can make the pipeline composition and optimisation methods to continue create invalid pipelines and can not find a valid pipeline within time budget. For example, AutoWeka4MCPS can find pipelines for the cases of the dataset convex and seed number 0 and 2, but we can not see any result using other seed numbers. 
Because AutoWeka4MCPS can compose pipelines which have up to six components, it is more likely to generate invalid pipelines and the evaluation time of these invalid pipelines are significant. For example, the wasted evaluation time is 97.98\% in the case of using the dataset car and Iter 5. We can see that changing the different random iterations has a strong impact on the wasted evaluation time in the case of AutoWeka4MCPS. For example, the experiments with the dataset abalone show that the wasted evaluation time is in the range between 0.66\% and 92.87\%. The reason is that Weka libraries them-self can evaluate the compatibility of a single component pipeline without execution. If the initialisation of the pipeline composition and optimisation with a specific seed number results in pipelines consisting of only one predictor, and these pipelines are well-performing, it tends to exploit similar ML pipelines. As a result, the wasted evaluation time is low.
However, this impact is negligible in the case of Auto-sklearn. The reason is that Auto-sklearn uses meta-learning to initialise with promising ML pipelines. %Therefore, the impact of changing seed numbers is negligible. 
The experiments with the datasets abalone, car and gcredit show that Auto-sklearn limits the generation of invalid pipelines by making assumption about cleaned input datasets, because the experiments crash if the input datasets have multiple attribute types. It means that Auto-sklearn can not handle invalid pipelines effectively.

\subsection{Experiments to compare the performance of AVATAR and the existing methods}

\begin{table}[htb!]

\vspace{-0.35cm}
\caption {Comparison of the performance of the AVATAR and T-method}
\label{tab:exp_avatar_tmethod}

\centering
\scriptsize

\begin{tabular}{|l|l|l|l|l|l|l|}
\hline
\multicolumn{2}{|l|}{\textbf{Dataset}}                                                                                                              & \textbf{abalone}                                             & \textbf{car}                                                 & \textbf{convex}                                             & \textbf{gcredit}                                             & \textbf{winequality}                                         \\ \hline
\multirow{2}{*}{\textbf{T-method}}   & \textbf{\begin{tabular}[c]{@{}l@{}}Invalid/ valid \\ pipelines\end{tabular}}                                 & \begin{tabular}[c]{@{}l@{}}683/\\ 1,097\end{tabular}         & \begin{tabular}[c]{@{}l@{}}4,387/\\ 6,817\end{tabular}       & \begin{tabular}[c]{@{}l@{}}252/\\ 428\end{tabular}          & \begin{tabular}[c]{@{}l@{}}4,557/\\ 7,208\end{tabular}       & \begin{tabular}[c]{@{}l@{}}1,276/\\ 1,951\end{tabular}       \\ \cline{2-7} 
                                     & \textbf{\begin{tabular}[c]{@{}l@{}}Total evaluation\\ time of invalid/\\ valid pipelines (s)\end{tabular}}   & \begin{tabular}[c]{@{}l@{}}27,711.9/\\ 15,484.1\end{tabular} & \begin{tabular}[c]{@{}l@{}}18,627.9/\\ 24,459.4\end{tabular} & \begin{tabular}[c]{@{}l@{}}5,818.3/\\ 37,765.1\end{tabular} & \begin{tabular}[c]{@{}l@{}}19,597.9/\\ 23,452.5\end{tabular} & \begin{tabular}[c]{@{}l@{}}10,830.1/\\ 32,326.9\end{tabular} \\ \hline
\multirow{2}{*}{\textbf{AVATAR}}     & \textbf{\begin{tabular}[c]{@{}l@{}}Invalid/ valid\\ pipelines\end{tabular}}                                  & \begin{tabular}[c]{@{}l@{}}663/\\ 1,117\end{tabular}         & \begin{tabular}[c]{@{}l@{}}4,387/\\ 6,817\end{tabular}       & \begin{tabular}[c]{@{}l@{}}250/\\ 430\end{tabular}          & \begin{tabular}[c]{@{}l@{}}4,552/\\ 7,213\end{tabular}       & \begin{tabular}[c]{@{}l@{}}1,262/\\ 1,965\end{tabular}       \\ \cline{2-7} 
                                     & \textbf{\begin{tabular}[c]{@{}l@{}}Total evaluation\\ time of invalid/\\ valid pipelines (s)\end{tabular}}   & 3.5/4.9                                                      & 43.1/64.8                                                    & 19.6/131.1                                                  & 57.0/89.2                                                    & 17.1/25.4                                                    \\ \hline
\multicolumn{2}{|l|}{\textbf{\begin{tabular}[c]{@{}l@{}}Pipelines have  different/similar\\ evaluated results\end{tabular}}}                        & 20/1,760                                                     & 0/11,204                                                     & 2/678                                                       & 5/11,760                                                     & 14/3,213                                                     \\ \hline
\multicolumn{2}{|l|}{\textbf{\begin{tabular}[c]{@{}l@{}}The percentage of\\ pipelines that the AVATAR\\ can validate accurately (\%)\end{tabular}}} & 98.88                                                        & 100.00                                                       & 99.71                                                       & 99.96                                                        & 99.57                                                        \\ \hline
\end{tabular}

\vspace{-0.35cm}

\end{table}

In order to demonstrate the efficiency of the AVATAR, we have conducted a second set of experiments. We run these experiments on a machine with an Intel core i7-8650U CPU and 16GB memory. We compare the performance of the AVATAR and the T-method that requires the executions of pipelines. The T-method is used to evaluate the validity of pipelines in the pipeline composition and optimisation of AutoWeka4MCPS and Auto-sklearn.
We randomly generate ML pipelines which have up to six components (i.e., these component types are missing value handling, dimensionality reduction, outlier removal, data transformation, data sampling and predictor). The predictor is put at the end of the pipelines because a valid pipeline always has a predictor at the end.
Each pipeline is evaluated by the AVATAR and the T-method. We set the time budget to 12 hours per dataset. We use the following criteria to compare the performance: the number of invalid/ valid pipelines, the total evaluation time of invalid/ valid pipelines (seconds), the number of pipelines that have the same evaluated results between the AVATAR and the T-method, and the percentage of the pipelines that the AVATAR can validate accurately (\%) in comparison to the T-method. 

Table \ref{tab:exp_avatar_tmethod} compares the performance of the AVATAR and the T-method using the above criteria. We can see that the total evaluation time of invalid/valid pipelines of the AVATAR is significantly lower than the T-method.
While the evaluation time of pipelines of the AVATAR is quite stable, the evaluation time of pipelines of the T-method is much higher and depends on the size of the datasets. It means that the AVATAR is faster than the T-method in evaluating both invalid and valid pipelines regardless of the size of datasets. Moreover, we can see that the accuracy of the AVATAR is approximately 99\% in comparison with the T-method. We have carefully reviewed the pipelines which have different evaluated results between the AVATAR and the T-method. Interestingly, the AVATAR evaluates all of these pipelines to be valid and vice versa in the case of the T-method. The reason is that executions of these pipelines cause the out of memory problem. In other words, the AVATAR does not consider the allocated memory as an impact on the validity of a pipeline.
%Obviously, if a pipeline requires more allocated memory to train a dataset, it's still an invalid pipeline.
A promising solution is to reduce the size of an input dataset by adding a sampling component with appropriate hyperparameters. If the sampling size is too small, we may miss important features. If the sampling size is large, we may continue to run into the problem of out of memory.  
We cannot conclude that if we allocate more memory, whether the executions of these pipelines would be successful or not. It proves that the validity of a pipeline also depends on its execution environment such as memory. These factors have not been considered yet in the AVATAR. This is an interesting research gap that should be addressed in the future.

\begin{table}[htb!]

\centering

\vspace{-0.35cm}
\caption {Five invalid pipelines with the longest evaluation time using the T-method on the gcredit dataset.}
\label{tab:top5_worst_pipeline}

\scriptsize

\begin{tabular}{|l|l|l|l|l|l|}
\hline
\textbf{Pipeline}     & \#1    & \#2    & \#3    & \#4    & \#5    \\ \hline
\textbf{T-method (s)} & 11.092 & 11.068 & 11.067 & 11.067 & 11.066 \\ \hline
\textbf{AVATAR (s)}   & 0.014  & 0.012  & 0.011  & 0.011  & 0.011  \\ \hline
\end{tabular}
\vspace{-0.35cm}

\end{table}

Finally, we take a detailed look at the invalid pipelines with the longest evaluation time using the T-method on the gcredit dataset, as shown in Table \ref{tab:top5_worst_pipeline}. Pipeline \#1 (11.092 s) has the structure \textit{ReplaceMissingValues} \(\rightarrow\) \textit{PeriodicSampling} \(\rightarrow\) \textit{NumericToNominal} \(\rightarrow\) \textit{PrincipalComponents} \(\rightarrow\) \textit{SMOreg}. This pipeline is invalid because \textit{SMOreg} does not work with nominal classes, and there is no component transforming the nominal to numeric data. We can see that the AVATAR is able to evaluate the validity of this pipeline without executing it in just 0.014 s.

%Although we tried to generate the synthetic datasets that have as many as possible transformed-features, the current version of the AVATAR knowledge base may not have enough transformed-features that lead to incorrect evaluations of pipelines.

\section{Conclusion}
\label{sec:conclusion}

We empirically demonstrate the problem of generation of invalid pipelines during pipeline composition and optimisation. We propose the AVATAR which is a pipeline evaluation method using a surrogate model. The AVATAR can be used to accelerate pipeline composition and optimisation methods by quickly ignoring invalid pipelines to improve the effectiveness of the AutoML optimisation process. In future, we will improve the AVATAR to evaluate pipelines' quality besides their validity. Moreover, we will investigate how to employ the AVATAR to reduce search spaces dynamically.
%Moreover, we will improve the AVATAR knowledge base to learn new transformed-features automatically during the pipeline composition and optimisation.
%We will also extend the AVATAR with the scikit-learn libraries to compare the performance of the search strategies such as random search and SMAC implementing the AVATAR for Auto-sklearn and AutoWeka4MCPS.

\section*{Acknowledgment}
This research is sponsored by AAi, University of Technology Sydney (UTS).

\bibliographystyle{splncs}
\bibliography{references}

\begin{thebibliography}{10}

\bibitem{kaga09}
Kadlec, P., Gabrys, B.:
\newblock Architecture for development of adaptive on-line prediction models.
\newblock Memetic Computing \textbf{1} (2009)  241

\bibitem{sabu18}
Salvador, M.M., Budka, M., Gabrys, B.:
\newblock Automatic composition and optimization of multicomponent predictive
  systems with an extended auto-weka.
\newblock IEEE Transactions on Automation Science and Engineering (2018)

\bibitem{zohu19}
Z{\"o}ller, M.A., Huber, M.F.:
\newblock Survey on automated machine learning.
\newblock arXiv preprint arXiv:1904.12054 (2019)

\bibitem{olmo16}
Olson, R.S., Moore, J.H.:
\newblock Tpot: A tree-based pipeline optimization tool for automating machine
  learning.
\newblock In: Workshop on Automatic Machine Learning. (2016)  66--74

\bibitem{fekl15}
Feurer, M., Klein, A., Eggensperger, K., Springenberg, J., Blum, M., Hutter,
  F.:
\newblock Efficient and robust automated machine learning.
\newblock In: Advances in Neural Information Processing Systems. (2015)
  2962--2970

\bibitem{mowe18}
Mohr, F., Wever, M., H{\"u}llermeier, E.:
\newblock Ml-plan: Automated machine learning via hierarchical planning.
\newblock Machine Learning \textbf{107} (2018)  1495--1515

\bibitem{giya18}
Gil, Y., Yao, K.T., Ratnakar, V., Garijo, D., Ver~Steeg, G., Szekely, P.,
  Brekelmans, R., Kejriwal, M., Luo, F., Huang, I.H.:
\newblock P4ml: A phased performance-based pipeline planner for automated
  machine learning.
\newblock In: AutoML Workshop at ICML. (2018)

\bibitem{depi17}
de~S{\'a}, A.G., Pinto, W.J.G., Oliveira, L.O.V., Pappa, G.L.:
\newblock Recipe: a grammar-based framework for automatically evolving
  classification pipelines.
\newblock In: European Conference on Genetic Programming, Springer (2017)
  246--261

\bibitem{tsga12}
Tsakonas, A., Gabrys, B.:
\newblock Gradient: Grammar-driven genetic programming framework for building
  multi-component, hierarchical predictive systems.
\newblock Expert Systems with Applications \textbf{39} (2012)  13253--13266

\bibitem{chmi09}
Chinosi, M., Trombetta, A.:
\newblock Modeling and validating bpmn diagrams.
\newblock In: 2009 IEEE Conference on Commerce and Enterprise Computing, IEEE
  (2009)  353--360

\bibitem{thhu13}
Thornton, C., Hutter, F., Hoos, H.H., Leyton-Brown, K.:
\newblock Auto-weka: Combined selection and hyperparameter optimization of
  classification algorithms.
\newblock In: Proceedings of the 19th ACM SIGKDD international conference on
  Knowledge discovery and data mining, ACM (2013)  847--855

\bibitem{sabu17b}
Salvador, M.M., Budka, M., Gabrys, B.:
\newblock Modelling multi-component predictive systems as petri nets.
\newblock (2017)

\bibitem{tafa10}
Tan, W., Fan, Y., Zhou, M., Tian, Z.:
\newblock Data-driven service composition in enterprise soa solutions: A petri
  net approach.
\newblock IEEE Transactions on Automation Science and Engineering \textbf{7}
  (2010)  686--694

\bibitem{sabu16b}
Salvador, M.M., Budka, M., Gabrys, B.:
\newblock Towards automatic composition of multicomponent predictive systems.
\newblock In: International conference on hybrid artificial intelligence
  systems, Springer (2016)  27--39

\bibitem{bava07}
Barker, A., Van~Hemert, J.:
\newblock Scientific workflow: a survey and research directions.
\newblock In: International Conference on Parallel Processing and Applied
  Mathematics, Springer (2007)  746--753

\bibitem{baal18}
Balaji, A., Allen, A.:
\newblock Benchmarking automatic machine learning frameworks.
\newblock arXiv preprint arXiv:1808.06492 (2018)

\end{thebibliography}

\end{document}